\documentclass[final]{cvpr}

\usepackage{times}
\usepackage{epsfig}
\usepackage{graphicx}
\usepackage{amsmath}
\usepackage{amssymb}

\usepackage{booktabs}
\usepackage{tabularx}
\usepackage{color}
\usepackage{colortbl}
\usepackage{subfigure}
\usepackage{multirow}
\usepackage{caption}

\usepackage[pagebackref=true,breaklinks=true,colorlinks,bookmarks=false]{hyperref}

\def\cvprPaperID{7856} 
\def\confYear{CVPR 2021}

\def\thickhline{%
  \noalign{\ifnum0=`}\fi\hrule \@height \thickarrayrulewidth \futurelet
   \reserved@a\@xthickhline}
\def\@xthickhline{\ifx\reserved@a\thickhline
               \vskip\doublerulesep
               \vskip-\thickarrayrulewidth
             \fi
      \ifnum0=`{\fi}}
\makeatother

\newlength{\thickarrayrulewidth}
\newcommand\blfootnote[1]{%
  \begingroup
  \renewcommand\thefootnote{}\footnote{#1}%
  \addtocounter{footnote}{-1}%
  \endgroup
}

\begin{document}

\title{Memory-guided Unsupervised Image-to-image Translation}

\author{Somi Jeong$^1$ \quad \quad Youngjung Kim$^2$ \quad \quad Eungbean Lee$^1$ \quad \quad Kwanghoon Sohn$^{1*}$\\
$^1$Department of Electrical \& Electronic Engineering, Yonsei University, Seoul, Korea\\
$^2$Agency for Defense Development (ADD), Daejeon, Korea\\
{\tt\small \{somijeong, eungbean, khsohn\}@yonsei.ac.kr, read12300@add.re.kr}}

\twocolumn[{%
  \renewcommand\twocolumn[1][]{#1}
  \maketitle   
  \vspace{-20pt}
  \begin{center}
	 \includegraphics[width=0.99\textwidth]{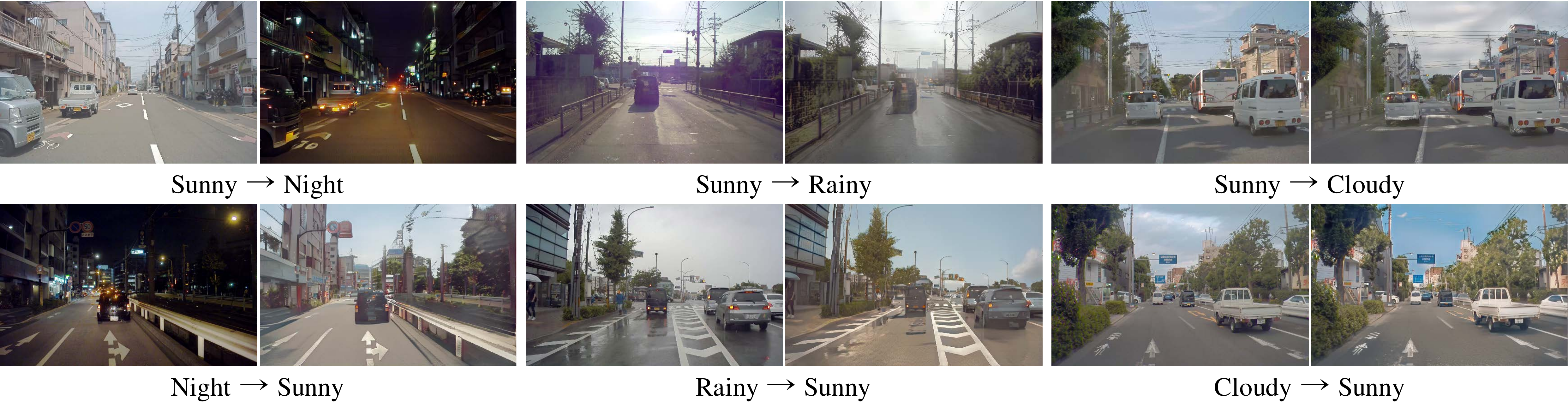}
	 \vspace{-7pt}
     \captionof{figure}{
     \textbf{Instance-level image-to-image translation.}
     We present a memory-guided unsupervised image-to-image translation method that performs diverse translation between two visual domains by leveraging a class-aware memory.}
     \label{fig:f1}
  \end{center}
}]

\blfootnote{
\hspace{-12pt}
This research was supported by the Agency for Defense Development under the grant UD2000008RD.

$^*$Corresponding author}
 
\vspace{-12pt}
\begin{abstract}
We present a novel unsupervised framework for instance-level image-to-image translation.
Although recent advances have been made by incorporating additional object annotations, existing methods often fail to handle images with multiple disparate objects.
The main cause is that, during inference, they apply a global style to the whole image and do not consider the large style discrepancy between instance and background, or within instances.
To address this problem, we propose a class-aware memory network that explicitly reasons about local style variations.
A key-values memory structure, with a set of read/update operations, is introduced to record class-wise style variations and access them without requiring an object detector at the test time.
The key stores a domain-agnostic content representation for allocating memory items, while the values encode domain-specific style representations.
We also present a feature contrastive loss to boost the discriminative power of memory items.
We show that by incorporating our memory, we can transfer class-aware and accurate style representations across domains.
Experimental results demonstrate that our model outperforms recent instance-level methods and achieves state-of-the-art performance.
\end{abstract}

\vspace{-10pt}
\section{Introduction}
\vspace{-4pt}
Unsupervised image-to-image (I2I) translation is the task of learning a mapping between unpaired images in diverse domains.
It can be applied to a variety of applications, including attribute manipulation~\cite{choi2018stargan,lee2020maskgan}, style transfer~\cite{ulyanov2017improved,huang2017arbitrary}, data augmentation~\cite{mariani2018bagan, huang2018auggan}, and domain adaptation~\cite{murez2018image,hoffman2018cycada}.
Recent methods~\cite{zhu2017unpaired,liu2017unsupervised,kim2017learning,taigman2017unsupervised,yi2017dualgan} achieved impressive results based on a cycle-consistency constraint that forces translated images to be mapped back to their original domain.
However, they usually assume a deterministic one-to-one mapping between two domains, thus failing to capture the full distribution of possible outputs.
Several methods~\cite{zhu2017toward,huang2018multimodal,lee2018diverse,gonzalez2018image,yang2019diversity} aim to model complex and multimodal distributions to generate diverse outputs.
They postulate that the image representation can be disentangled into domain-invariant content and domain-specific style.
However, they simply formulate I2I translation as a global translation problem and apply a global content/style to entire images, which is problematic when handling complex images with many disparate objects.
Recently, INIT \cite{shen2019towards} and DUNIT \cite{bhattacharjee2020dunit} alleviated this problem by separately treating object instances and background with additional object annotations.
During training, INIT~\cite{shen2019towards} independently translates the instances using a separate reconstruction loss along with the global translation module.
At test time, however, it only uses the global module and discards the instance-level information.
DUNIT~\cite{bhattacharjee2020dunit} integrates an object detector within the I2I translation module and adds an instance-level encoder to extract instance-boosted features.
Although it can leverage the object instances at test time, it is not flexible enough to model diverse local style variations.
Furthermore, both methods require an off-the-shelf computationally expensive object detection module at test time.

Motivated by the aforementioned problems, in this paper, we introduce a novel instance-level I2I translation framework with an external memory module.
Specifically, we propose a class-aware memory network that can accurately store and propagate local-style information across different visual domains.
It comprises several class-wise memory matrices, and each matrix contains a set of key-values (items).
The key is used to address relevant memory items with respect to queries, and covers a shared content space.
Conversely, the values encode domain-specific style representations for its paired key.
This memory module allows storing diverse styles for different object instances into memory items during training (\emph{update}) and efficiently accessing them without an explicit object detector at test time (\emph{read}). 
Furthermore, we present a feature contrastive loss to enhance the discriminative power of memory items.
We show that, by incorporating our memory, the proposed method can capture the object details and reconstruct realistic images.
Experimental results on standard benchmarks, including INIT~\cite{shen2019towards}, KITTI~\cite{geiger2012we}, and Cityscapes~\cite{cordts2016cityscapes}, demonstrate the effectiveness of our method, which outperforms state-of-the-art instance-level I2I translation methods.
Furthermore, we demonstrate that our approach can be applied to domain adaptation detection tasks.

Our contributions can be summarized as follows:
\vspace{-5pt}
\begin{itemize}	
	\item We propose a memory-guided unsupervised I2I translation (MGUIT) framework that stores and propagates instance-level style information across visual domains.
	      To best of our knowledge, this is the first work that explores a memory network in I2I translation.
 	\vspace{-6pt}
 	\item We introduce a key-values memory structure to effectively record diverse style variations and access them during I2I translation. 
 	      Our model does not require explicit object detection modules at test time.
 	      We also propose a feature contrastive loss to improve the diversity and discriminative power of our memory items.
 	\vspace{-6pt}
 	\item Our method produces realistic translation results while preserving instance details well; it outperforms recent state-of-the-art methods on standard benchmarks.
\end{itemize}

\vspace{-10pt}
\section{Related Work}
\vspace{-5pt}
\paragraph{Image-to-image translation.}
The seminal work of Pix2Pix~\cite{isola2017image} achieved impressive results in I2I translation tasks using paired images based on conditional generative adversarial networks (GANs)~\cite{mirza2014conditional}.
To reduce the difficulty in collecting the image pairs, various unsupervised I2I translation approaches~\cite{zhu2017unpaired,liu2017unsupervised,kim2017learning,taigman2017unsupervised,yi2017dualgan} have been proposed.
They mainly regularize ill-posed training procedure by adopting a cycle consistency constraint, which enforces the translated image from source to target domain to be mapped back to the source domain.
Because they model a deterministic one-to-one mapping, they failed to generate diverse outputs.
To tackle this limitation, some methods have extended it into multi-modal/multi-domain mapping~\cite{zhu2017toward,huang2018multimodal,lee2018diverse,gonzalez2018image,yang2019diversity}.
Based on the assumption that images can be disentangled into shared content and separate style representations, they apply various learning strategies to enhance their generalization capabilities, such as weight sharing~\cite{lee2018diverse, huang2018multimodal}, variational autoencoder~\cite{huang2018multimodal, gonzalez2018image}, and normalization layer~\cite{ulyanov2017improved,dumoulin2016learned,huang2017arbitrary}.
Unfortunately, they show poor results when translating images with multiple instances because they do not consider instance-level information.

\begin{figure*}[ht]
	\centering
	{\includegraphics[width=0.95\linewidth]{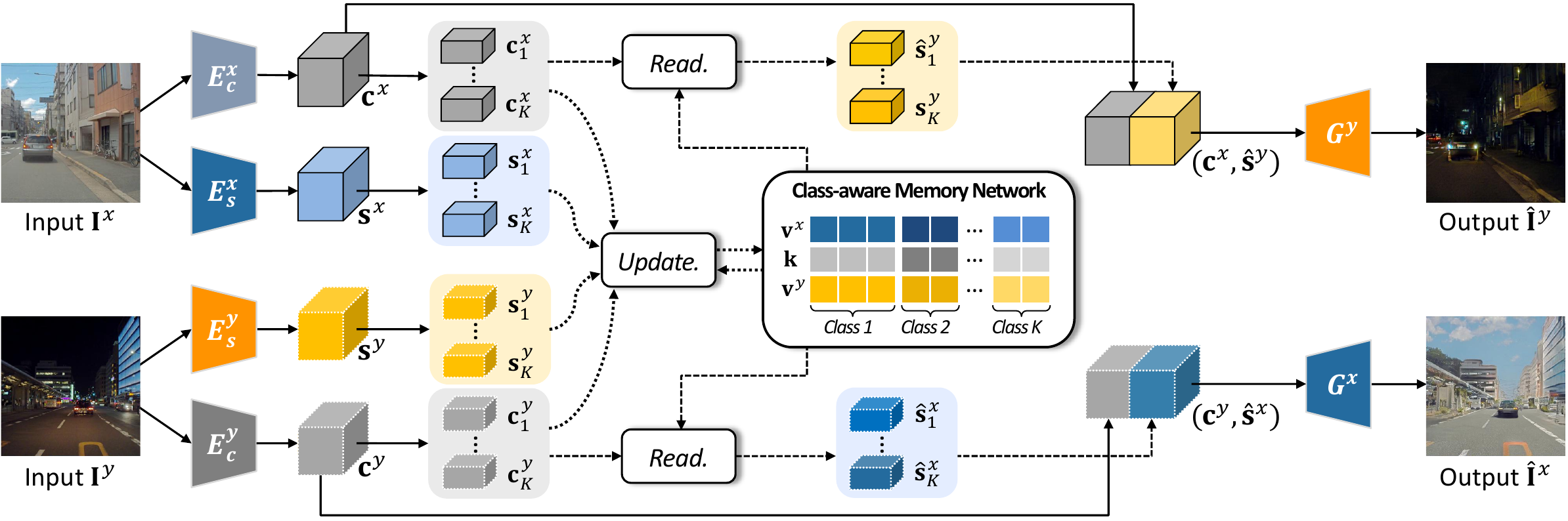}}\vspace{-2pt}
	\caption{
		\textbf{The overview of the proposed architecture.}
		The content and style encoders extract content $\mathbf{c}^x$ and style $\mathbf{s}^x$ features from the input image $\mathbf{I}^x$ and they are clustered by object class $\{(\mathbf{c}^x_1, \mathbf{s}^x_1), \cdots, (\mathbf{c}^x_K, \mathbf{s}^x_K)\}$.
        The class-aware memory network consists of key-values memory items $(\mathbf{k}, \mathbf{v}^x, \mathbf{v}^y)$ assigned to each object class and uses $(\mathbf{c}^x_k, \mathbf{s}^x_k)$ to read and update memory items.
		The generator takes the enhanced style feature maps $\hat{\mathbf{s}}^y$ retrieved from memory and generates the image $\hat{\mathbf{I}}^y$.
		}
	\vspace{-10pt}
	\label{fig:f2}
\end{figure*}

\vspace{-12pt}
\paragraph{Instance-level image-to-image translation}
Very recently, several efforts have been dedicated to achieving instance-level I2I translation~\cite{mo2019instagan,shen2019towards,bhattacharjee2020dunit}.
InstaGAN~\cite{mo2019instagan} performs the instance-level image translation using the object segmentation masks as extra supervision while maintaining the background.
On the other hand, INIT~\cite{shen2019towards} and DUNIT~\cite{bhattacharjee2020dunit} focus on translating instances and backgrounds simultaneously, which is the same objective as our work.
INIT~\cite{shen2019towards} employs the instance and global styles separately to guide the generation of target domain objects directly.
In inference, however, it uses the global style only, thus neglecting the instance style.
DUNIT~\cite{bhattacharjee2020dunit} incorporates the object detector and I2I translation to extract the instance-boost feature representations.
Since the global and instance features are unified using a global style, its translated results may lose the inherent instance characteristic.
Different from the aforementioned methods, we aim to infer the instance style in both training and testing time to produce more realistic results.
To this end, we adopt the novel memory networks, which store the style information during training and read the appropriate style representation for inference.

\vspace{-12pt}
\paragraph{Memory networks.}
Memory network~\cite{weston2015memory,sukhbaatar2015end} is a learnable neural network module, which stores information in external memory and reads the relevant contents from the memory.
The Key-Value Memory Networks~\cite{miller2016key} was introduced, which exploits a key-value structured memory for reading documents.
Given a query, the key is used to retrieve relevant memories, and its corresponding values are returned.
Thanks to its high flexibility that it records different knowledge in the key and value, it has been widely adopted in solving various vision problems such as natural language
processing~\cite{kumar2016ask,daniluk2017frustratingly}, movie understanding~\cite{na2017read}, visual tracking~\cite{yang2018learning}, and video object segmentation~\cite{oh2019video,miao2020memory}.

Inspired by \cite{miller2016key}, we introduce a key-values structured memory, modified to be suitable for I2I translation.
Recently, DM-GAN~\cite{zhu2019dm} adopts a dynamic memory network to generate a high-quality image from text descriptions.
They select the relevant value by comparing the key memory with the input text, and it is used to generate the image.
In contrast, we employ the key-values memory to store domain-agnostic content representations and domain-specific style representations.

\vspace{-3pt}
\section{Proposed Method}
\vspace{-3pt}
We denote by $\mathcal{X}$ and $\mathcal{Y}$ two visual domains, e.g., sunny and night (or rainy).
Our objective is to learn a multi-modal mapping between $\mathcal{X}$ and $\mathcal{Y}$ by accurately storing and propagating class-aware style information.
To this end, we introduce a novel memory network along with an I2I network to explicitly explain the objects.
The memory network contains several memory items; each memory item stores class-aware feature representations. 
The features from the I2I encoders, i.e., \emph{queries}, are used to \emph{read} and \emph{update} class-aware features in the memory.
The I2I generator then inputs them to reconstruct the final translated image.
An overview of our framework is illustrated in Fig. \ref{fig:f2}.
We assume that, during training time, we can access the ground-truth object annotations (bounding box and class) to update the memory items assigned for each class.
At test time, however, no object annotations are required given that we can retrieve the appropriate memory items through the read operations.
Next, we comprehensively describe the components of the MGUIT framework.

\vspace{-3pt}
\subsection{Image-to-Image Translation Network}
\vspace{-3pt}
We basically follow the DRIT~\cite{lee2018diverse} architecture\footnote{We thus omit unnecessary details to avoid repetition.}.
Our architecture consists of two coupled content encoders $E_c=\{E^x_c, E^y_c\}$, style encoders $E_s=\{E^x_s, E^y_s\}$, and generators $\{G^x, G^y\}$ in each domain, $\mathcal{X}$ or $\mathcal{Y}$.
For adversarial learning, it additionally contains domain discriminators $\{D^x, D^y\}$ to determine whether the image is from its original domain, and a content discriminator $D_c$.
As in \cite{lee2018diverse}, we decompose an image $\mathbf{I}$ into a domain-agnostic content space $\mathbf{c} \in \mathcal{C}$ and a domain-specific style space $\mathbf{s} \in \mathcal{S}$, where $(\mathbf{c}^{x}, \mathbf{c}^{y}) = (E_c^{x}(\mathbf{I}^x), E_c^{y}(\mathbf{I}^y))$ and $(\mathbf{s}^{x}, \mathbf{s}^{y}) = (E_s^{x}(\mathbf{I}^x), E_s^{y}(\mathbf{I}^y))$.
The existing I2I methods \cite{lee2018diverse, huang2018multimodal, gonzalez2018image} simply swap $\mathbf{s}$ from both domains ($\mathcal{X} \leftrightarrow \mathcal{Y}$) to produce $\hat{\mathbf{I}}^{y} = G^y(\mathbf{c}^x, \mathbf{s}^y)$ (and vice versa for $\hat{\mathbf{I}}^{x}$). 
This strategy performs a global-style translation over the entire image, making the results for complex scenes with multiple objects less realistic. 
In contrast, we use an external class-aware memory network $M$ that records diverse intra- and inter-class style variations simultaneously.
Through a read operation, the memory $M$ takes $\mathbf{c}$ as query maps and outputs the enhanced style feature maps $\hat{\mathbf{s}}$.
Finally, the generators reconstruct the translated images by combining $\mathbf{c}$ and $\hat{\mathbf{s}}$ as:
\begin{equation}\label{eq1}
\hat{\mathbf{I}}^x=G^x(\mathbf{c}^y,\hat{\mathbf{s}}^{x}),~~~~
\hat{\mathbf{I}}^y=G^y(\mathbf{c}^x,\hat{\mathbf{s}}^{y}).
\end{equation}
Next, we describe how to read the appropriate style and update $M$ according to the object classes.  

\vspace{-2pt}
\subsection{Class-aware Memory Network}\vspace{-4pt}
The memory network contains $N$ memory items to store class-aware feature representations.
We assign $N_k$ items to each class, where $\Sigma_{k=1}^{K}N_k = N$ and $K$ is the total number of classes (including the background).
$N_k$ is the parameter used to model the intra-class variation, which can vary according to the class. 
For example, we can assign $4$ and $6$ memory items to ``\emph{car}" and ``\emph{background}" classes, respectively, for a total of $N=10$ memory items. 
Each item consists of a pair of $1 \times 1 \times C$ vectors $(\mathbf{k}, \mathbf{v}^x, \mathbf{v}^y)$, where $C$ is the number of channels.
$\mathbf{k}$ denotes the shared \textbf{key} used to address items, and also encodes the domain-agnostic content representations.
Similarly, \textbf{values} $(\mathbf{v}^x, \mathbf{v}^y)$ store domain-specific style representations for the paired $\mathbf{k}$.
This \textbf{key}-\textbf{values} memory structure allows recording diverse style variations into memory items and accessing them during I2I translation without an off-the-shelf object detector.

\begin{figure}
	\centering
	\includegraphics[width=0.9\linewidth]{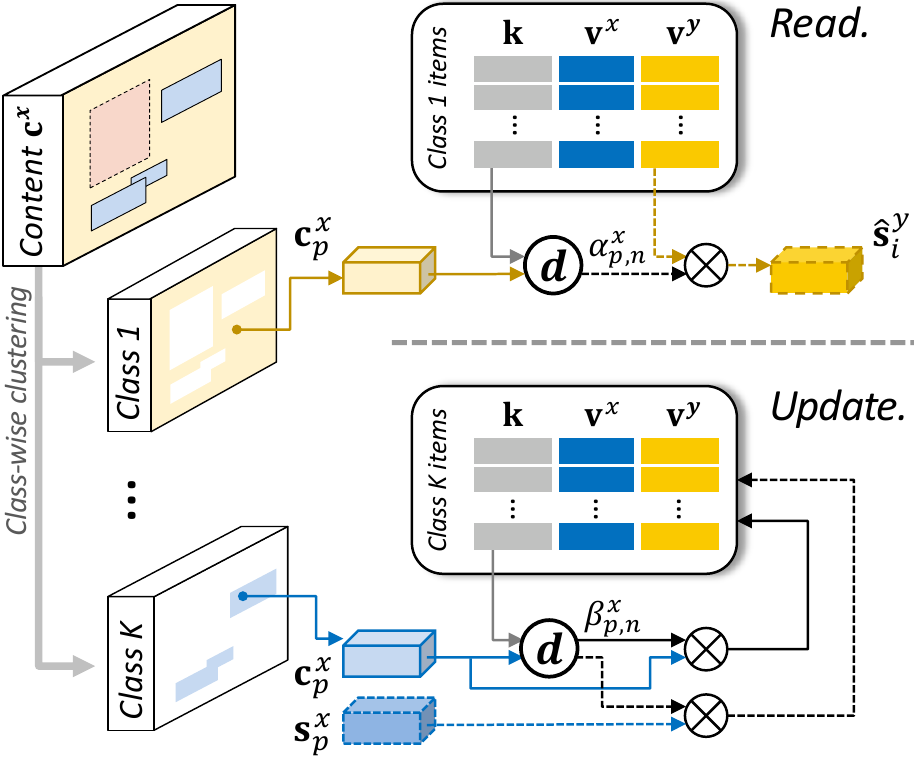}\vspace{-4pt}
	\caption{\textbf{Read and update operations for training.}
	We cluster features by class, and the read and update operations are processed class-wisely.
For read, we compute a read-weight $\alpha^x_{p,n}$ between each $\mathbf{c}^x_p$ and all memory keys $\mathbf{k}$ in (\ref{eq2}).
The aggregated style feature $\hat{\mathbf{s}}^y_p$ is retrieved by taking a weighted average of the cross-domain values as (\ref{eq4}).
For update, we compute an update-weight $\beta^x_{p,n}$ in (\ref{eq5}), and update the key and values as (\ref{equ:write}).
(and vice versa for domain $\mathcal{Y}$)}
	\vspace{-5pt}
	\label{fig:f3}
\end{figure}

Given the object annotations, we first cluster ($\mathbf{c}, \mathbf{s}$) into a set of features $\{(\mathbf{c}_1, \mathbf{s}_1), \cdots, (\mathbf{c}_K, \mathbf{s}_K)\}$ to train the memory network.
We feed the class-wise cluster $(\mathbf{c}_k, \mathbf{s}_k)$ to only read/write the corresponding $N_k$ memory items, as shown in Fig.~\ref{fig:f3}.
Next, the subscript $k$ is omitted for simplicity, but we note that $(\mathbf{c}_k, \mathbf{s}_k)$ are only applied to the corresponding $N_k$ items assigned to class $k$.

\vspace{-8pt}
\paragraph{Read.}
To read the appropriate style values, we compute the similarity between each $\mathbf{c}_p$ and $\mathbf{k}$, resulting in a read-weight matrix $\alpha^{x \ (\text{or} \ y)} \in \mathbb{R}^{P \times N}$ :
\begin{equation}\label{eq2}
\resizebox{0.45\textwidth}{!}{$
\alpha_{p, n}^{x} = \frac{\exp(d(\mathbf{c}_{p}^x, \mathbf{k}_n))}{\sum_{n^{'}=1}^{N} \exp(d(\mathbf{c}_p^x, \mathbf{k}_{n^{'}}))},
\alpha_{p, n}^{y} = \frac{\exp(d(\mathbf{c}_{p}^y, \mathbf{k}_n))}{\sum_{n^{'}=1}^{N} \exp(d(\mathbf{c}_p^y, \mathbf{k}_{n^{'}}))},$}
\end{equation}
where $\mathbf{c}_{p}$ denotes individual features ($p = 1, \cdots, P$) of size $1 \times 1 \times C$, and $P$ is the total number of pixels in $\mathbf{c}$.
$d(\cdot,\cdot)$ is defined using cosine similarity as follows:
\begin{equation}
d(\mathbf{c}_p, \mathbf{k}_n) = \frac{\mathbf{c}_p\mathbf{k}^\top_n}{\left \|\mathbf{c}_p\right \|_2 \left \|\mathbf{k}_n\right \|_2}.
\end{equation}
Inspired by \cite{miller2016key}, we read the memory item by taking a weighted average of the cross-domain values:
\begin{equation} \label{eq4}
\hat{\mathbf{s}}^{x}_{p} =\textstyle \sum \nolimits_{n^{'}=1}^{N}\alpha^y_{p, n^{'}}\mathbf{v}^{x}_{n^{'}},~~~~
\hat{\mathbf{s}}^{y}_{p} = \sum \nolimits_{n^{'}=1}^N \alpha^x_{p, n^{'}}\mathbf{v}^{y}_{n^{'}}.
\end{equation}
This step is repeated for all $\mathbf{c}_{p}^{x \ (\text{or} \ y)}$, and produces an enhanced and aggregated style feature map $\hat{\mathbf{s}}^{y \ (\text{or} \ x)}$\footnote{Specifically, $(\mathbf{s}_1, \cdots, \mathbf{s}_k)$ are separately processed with the corresponding $N_k$ memory items assigned to class $k$ (see Fig. \ref{fig:f3}), and then merged into $\hat{\mathbf{s}}$ using their original coordinates in $\mathbf{s}$.}.
Through (\ref{eq4}), our model can transfer class-aware and spatially varying style information across domains ($\mathcal{X} \leftrightarrow \mathcal{Y}$) by referring to their content characteristics.
The translated images ($\hat{\mathbf{I}}^x, \hat{\mathbf{I}}^y$) can be obtained according to (\ref{eq1}).

\vspace{-8pt}
\paragraph{Update.}
To enrich the memory items, we also select and store class-aware features into the memory while removing redundant features from the memory.
Similar to the read operation, we calculate an update-weight matrix $\beta^{x \ (\text{or} \ y)} \in \mathbb{R}^{P \times N}$ between $\mathbf{c}$ and $\mathbf{k}$:
\begin{equation}\label{eq5}
\resizebox{0.45\textwidth}{!}{$
\beta_{p, n}^{x} = \frac{\exp(d(\mathbf{c}_p^{x}, \mathbf{k}_n))}{\sum_{p^{'}=1}^P \exp(d(\mathbf{c}^{x}_{p^{'}}, \mathbf{k}_{n}))},
\beta_{p, n}^{y} = \frac{\exp(d(\mathbf{c}_p^{y}, \mathbf{k}_n))}{\sum_{p^{'}=1}^P \exp(d(\mathbf{c}^{y}_{p^{'}}, \mathbf{k}_{n}))},
$}
\end{equation}
where we apply the softmax function along the $\mathbf{c}$-direction, as opposed to (\ref{eq2}). 
The update-weight matrix $\beta$ is used to assign the extracted content $\mathbf{c}$ and style features $\mathbf{s}$ to the relevant memory item.
The items $(\mathbf{k}_n, \mathbf{v}^x_n, \mathbf{v}^y_n)$ are updated using ($\mathbf{c}_p, \mathbf{s}_p$) weighted by $\beta$ as follows:
\begin{equation}\label{equ:write}
\begin{split}
&\hat{\mathbf{k}}_n=\small\|\mathbf{k}_n+\textstyle \sum \nolimits_{p^{'}=1}^P \beta_{p^{'},n}^{x}\mathbf{c}^x_{p^{'}}+ \sum \nolimits_{p^{'}=1}^P \beta_{p^{'},n}^{y}\mathbf{c}^y_{p^{'}}\small \| _2,\\[3pt]
&\hat{\mathbf{v}}^x_n=\small \|\mathbf{v}^x_n + \textstyle \sum \nolimits_{p^{'}=1}^P \beta_{p^{'},n}^{x}\mathbf{s}^x_{p^{'}}\small \|_2, \\[3pt]
&\hat{\mathbf{v}}^y_n=\small \|\mathbf{v}^y_n + \textstyle \sum \nolimits_{p^{'}=1}^P \beta_{p^{'},n}^{y}\mathbf{s}^y_{p^{'}}\small \|_2.
\end{split}
\end{equation}
We utilize both ($\mathbf{c}^x_p,\mathbf{c}^y_p)$ to update $\mathbf{k}_n$ because it records the shared content representations.
In contrast, the domain-specific values ($\mathbf{v}^x_n,\mathbf{v}^y_n)$ are updated individually.
We train the memory with a large number of images and ground-truth object annotations, thus enabling the most representative and discriminative features to be stored.

At test time, we compute $\alpha$ for all memory keys $\mathbf{k}$ without considering class information and retrieve the style values using (\ref{eq2}) and (\ref{eq4}).
We find that this strategy still works well because our memory is discriminatively trained using ground-truth object annotations.

\subsection{Loss Functions}
\vspace{-3pt}
\subsubsection{Image-to-image translation network }
\vspace{-3pt}
Following DRIT~\cite{lee2018diverse}, we adopt several loss functions to facilitate proper image reconstruction as follows.
\vspace{-10pt}
\paragraph{Reconstruction loss} makes the translated image similar to its original image~\cite{zhu2017unpaired,liu2017unsupervised}, which regularizes the ill-posed unsupervised I2I translation problem.
It consists of two terms, namely self-reconstruction $\mathcal{L}^{self}$ and cycle-reconstruction $\mathcal{L}^{cyc}$, which are expressed as 
\begin{equation}\label{equ:recloss}
\resizebox{0.43\textwidth}{!}{$
\begin{split}
\mathcal{L}^{self}&=\mathbb{E}_{x,y}[\left\|G^x(\mathbf{c}^x,\hat{\mathbf{s}}^{x})-\mathbf{I}^x\right\|_1 + \left\|G^y(\mathbf{c}^y, \hat{\mathbf{s}}^{y})-\mathbf{I}^y\right\|_1],\\
\mathcal{L}^{cyc}&=\mathbb{E}_{x,y}[\left\|G^x(\hat{\mathbf{c}}^y,\hat{\mathbf{s}}^{x})-\mathbf{I}^x\right\|_1 + \left\|G^y(\hat{\mathbf{c}}^x, \hat{\mathbf{s}}^{y})-\mathbf{I}^y\right\|_1],
\end{split}
$}\end{equation}
where $(\hat{\mathbf{c}}^x, \hat{\mathbf{c}}^y)$ denotes the content features from $(\hat{\mathbf{I}}^x, \hat{\mathbf{I}}^y)$.
\vspace{-10pt}
\paragraph{Adversarial loss} aims to minimize the distribution discrepancy between two different features, widely used in GANs~\cite{goodfellow2014generative,mirza2014conditional}.
We adopt two adversarial loss functions:
the content adversarial loss $\mathcal{L}^{adv}_c$ between $\mathbf{c}^x$ and $\mathbf{c}^y$, and the domain adversarial loss $\mathcal{L}^{adv}_d$ between $\mathcal{X}$ and $\mathcal{Y}$.
\vspace{-12pt}
\paragraph{KL loss} $\mathcal{L}^{KL}$ makes the style representation to be close to a prior Gaussian distribution.
\vspace{-12pt}
\paragraph{Latent regression loss} $\mathcal{L}^{latent}$ enforces the mappings between the style and the image to be invertible.

\vspace{-5pt}
\subsubsection{Class-aware memory network }
\vspace{-5pt}
It is important to store representative and discriminative class-aware features in the memory.
To this end, we propose a feature contrastive loss function.
\vspace{-12pt}
\paragraph{Feature contrastive loss} 
For each feature $\mathbf{c}_p$ (or $\mathbf{s}_p$), we define its nearest item $\mathbf{k}_{p+}$ (or $\mathbf{v}_{p+}$) as a positive sample, and the others as negative samples.
The distances to the positive/negative samples are penalized as follows:
\begin{equation}\label{equ:conloss}
\begin{split}
\mathcal{L}_{\mathbf{k}}^{con}&=-\sum_{p=1}^P\log \frac{\exp(\mathbf{c}_p \cdot \mathbf{k}_{p+} / \tau)}{\sum_{n=1}^N \exp(\mathbf{c}_p \cdot \mathbf{k}_{n} / \tau)},\\
\mathcal{L}_{\mathbf{v}}^{con}&=-\sum_{p=1}^P\log \frac{\exp(\mathbf{s}_p \cdot \mathbf{v}_{p+} / \tau)}{\sum_{n=1}^N \exp(\mathbf{s}_p \cdot \mathbf{v}_{n} / \tau)},
\end{split}
\end{equation}
for both domains, $\mathcal{X}$ and $\mathcal{Y}$. $\tau$ is a temperature parameter that controls the distribution concentration level. 

This is conceptually similar to feature separateness loss in \cite{park2020learning}, which encourages the queries to be close to the nearest item and separates individual items in the memory.
However, they only consider the second-nearest item as a negative sample using triplet loss~\cite{simo2015discriminative}.
Thus, the selection method of the second-nearest item has a high impact on the training efficiency and final performance.
By contrast, the proposed feature contrastive loss compares all items in the memory.
It is more effective for learning good feature representations and clustering in an unsupervised manner.

As a summary, the full objective function is as follows:
\begin{equation}\label{equ:full}
\begin{split}
\min_{E_c, E_s, G} &~\max_{D, D_c} ~\lambda^{self}\mathcal{L}^{self}+\lambda^{cyc}\mathcal{L}^{cyc}+\lambda^{adv}_{c}\mathcal{L}^{adv}_{c}\\
&+\lambda^{adv}_{d}\mathcal{L}^{adv}_{d}+\lambda^{KL}\mathcal{L}^{KL}+\lambda^{latent}\mathcal{L}^{latent}\\
&+\lambda_{\mathbf{k}}^{con}\mathcal{L}_{\mathbf{k}}^{con}+\lambda_{\mathbf{v}}^{con}\mathcal{L}_{\mathbf{v}}^{con},
\end{split}
\end{equation}
where the $\lambda$s control the importance of each term.

\vspace{-5pt}
\section{Experiments}
\vspace{-3pt}
\subsection{Experimental Settings}
\vspace{-3pt}
\paragraph{Implementation Details.}
Our networks were implemented based on DRIT\footnote{https://github.com/HsinYingLee/DRIT} with PyTorch~\cite{paszke2017automatic} and trained on one single NVIDIA TITAN RTX GPU.
Every network weights of each layer are initialized by a Gaussian distribution with a zero mean and a standard deviation of 0.001.
The Adam solver~\cite{kingma2015adam} was employed for optimization, where $\beta_1=0.9$, $\beta_2=0.999$.
The batch size was set to 1.
The initial learning rate was set to 0.0001 and 1, kept for first 30 epochs, and linearly decayed to zero over the next 30 epochs.
We set the number of memory items as 20 and its channel $C$ as 256.
We resize the short side of images to 360 pixels and crop it to $360 \times 360$ to train our framework.
The hyperparameters $\{\lambda^{self}, \lambda^{cyc}, \lambda^{adv}_{c}, \lambda^{adv}_{d}, \lambda^{KL}, \lambda^{latent}\}$ for I2I translation network are set the same as DRIT~\cite{zhu2017toward}, and $\{\lambda_{k}^{con}, \lambda_{v}^{con}\}$ are empirically determined 1 and 0.5.
Our code will be made publicly available.
\begin{figure*}[] 
	\centering
	\renewcommand{\thesubfigure}{}
	{\includegraphics[width=0.16\linewidth]{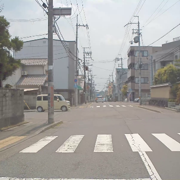}}
	{\includegraphics[width=0.16\linewidth]{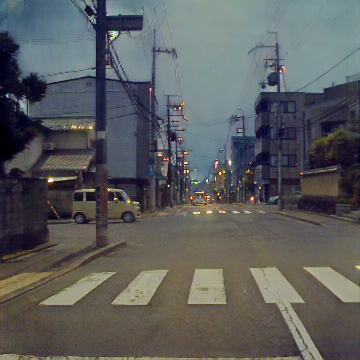}}
 	{\includegraphics[width=0.16\linewidth]{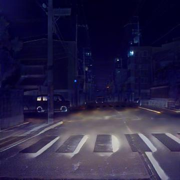}}
	{\includegraphics[width=0.16\linewidth]{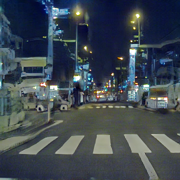}}
	{\includegraphics[width=0.16\linewidth]{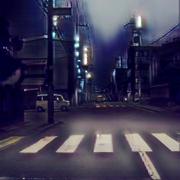}}
	{\includegraphics[width=0.16\linewidth]{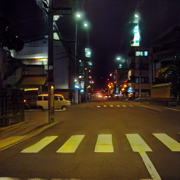}} \\
    {\includegraphics[width=0.16\linewidth]{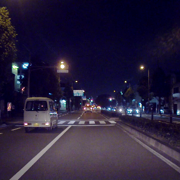}}
	{\includegraphics[width=0.16\linewidth]{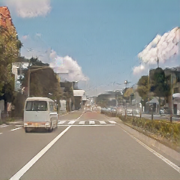}}
 	{\includegraphics[width=0.16\linewidth]{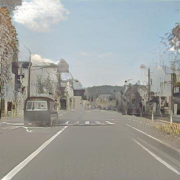}}
	{\includegraphics[width=0.16\linewidth]{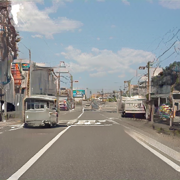}}
	{\includegraphics[width=0.16\linewidth]{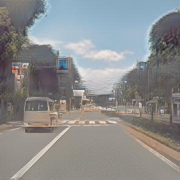}}
	{\includegraphics[width=0.16\linewidth]{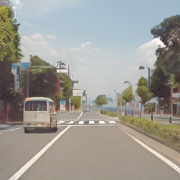}} \\
 	{\includegraphics[width=0.16\linewidth]{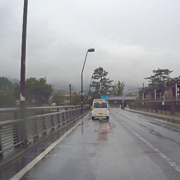}}
	{\includegraphics[width=0.16\linewidth]{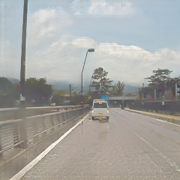}}
 	{\includegraphics[width=0.16\linewidth]{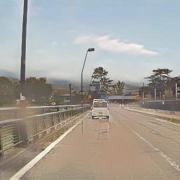}}
	{\includegraphics[width=0.16\linewidth]{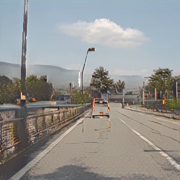}}
	{\includegraphics[width=0.16\linewidth]{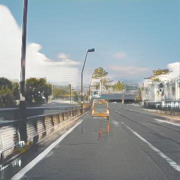}}
	{\includegraphics[width=0.16\linewidth]{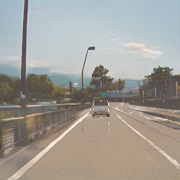}} \\\vspace{-5pt}
    \subfigure[(a) Input]
	{\includegraphics[width=0.16\linewidth]{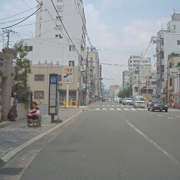}}
	\subfigure[(b) CycleGAN~\cite{zhu2017unpaired}]
	{\includegraphics[width=0.16\linewidth]{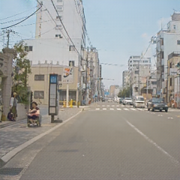}}
 	\subfigure[(c) UNIT~\cite{huang2018multimodal}]
 	{\includegraphics[width=0.16\linewidth]{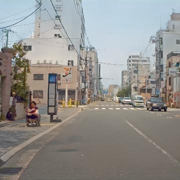}} 
    \subfigure[(d) MUNIT~\cite{huang2018multimodal}]
	{\includegraphics[width=0.16\linewidth]{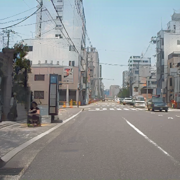}}
	\subfigure[(e) DRIT~\cite{lee2018diverse}]
	{\includegraphics[width=0.16\linewidth]{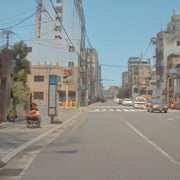}}
	\subfigure[(f) Ours]
	{\includegraphics[width=0.16\linewidth]{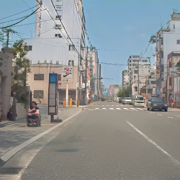}} \\ \vspace{-10pt}
	\caption{\textbf{Qualitative comparison of existing I2I translation methods.}
	(Top to bottom) sunny$\rightarrow$night, night$\rightarrow$sunny, rainy$\rightarrow$sunny, and cloudy$\rightarrow$sunny results.
    Our results preserve object details well and look realistic. Best viewed in color.
    }\vspace{-10pt}
	\label{fig:f4}
\end{figure*}

\vspace{-10pt}
\paragraph{Datasets.}
We conduct experiments on three datasets.

(1) \emph{INIT dataset}~\cite{shen2019towards} consists of 155K street scene images, including 4 domain categories (sunny, night, rainy, and cloudy).
It provides instance bounding box and object class annotations for car, person, and traffic sign.
We set the number of memory items for each class as 5, 3, and 2, and for the background as 10.
Following INIT~\cite{shen2019towards}, we use 85$\%$ images for training and 15$\%$ images for testing.
We conduct three translation experiments for sunny$\leftrightarrow$night, sunny$\leftrightarrow$rainy, and sunny$\leftrightarrow$cloudy.

\vspace{3pt}
(2) \emph{KITTI object detection benchmark}~\cite{geiger2012we} and \emph{Cityscapes dataset}~\cite{cordts2016cityscapes} are used to demonstrate that our method can help with domain adaptation.
KITTI benchmark~\cite{geiger2012we} contains 7,481 images for training and 7,518 images for testing, and it provides the bounding boxes for 6 object classes.
Cityscapes dataset~\cite{cordts2016cityscapes} is widely exploited for semantic segmentation,
which consists of 5,000 images with pixel-level annotations for 30 classes.
These datasets are used to conduct the domain adaptation for object detection (KITTI $\rightarrow$ Cityscapes case).
To integrate two datasets'  the object classes, we set the common 4 object classes as person, car, truck, and bicycle. 
We build 3 memory items for each class, including 8 background memory items.

\vspace{-10pt}
\paragraph{Compared methods.} 
We perform the evaluation on the following methods.
\vspace{-5pt}
\begin{itemize}
\item CycleGAN~\cite{zhu2017unpaired} and UNIT~\cite{liu2017unsupervised} are the typical unsupervised I2I translation methods.
\vspace{-5pt}
\item MUNIT~\cite{huang2018multimodal} and DRIT~\cite{lee2018diverse} are the multi-modal unsupervised I2I translation methods that are extensions of CycleGAN~\cite{zhu2017unpaired} and UNIT~\cite{liu2017unsupervised}. Especially, we exploit DRIT~\cite{lee2018diverse} as our baseline model.
\vspace{-5pt}
\item INIT~\cite{shen2019towards} and DUNIT~\cite{bhattacharjee2020dunit} are the existing instance-level unsupervised I2I methods. These methods are compared only for quantitative evaluation and not included in the qualitative comparison, since their code (parameters) is not publicly available.
\end{itemize}

\vspace{-15pt}
\paragraph{Evaluation metrics.}
Following the experimental protocol of existing unsupervised I2I translation methods, we evaluate our methods with Inception Score (IS)~\cite{salimans2016improved}, Conditional Inception Score (CIS)~\cite{huang2018multimodal}, and LPIPS Metric~\cite{zhang2018unreasonable}.

\subsection{Comparison to state-of-the-art}
\begin{figure} 
\centering
\vspace{10pt}
\includegraphics[width=1\linewidth]{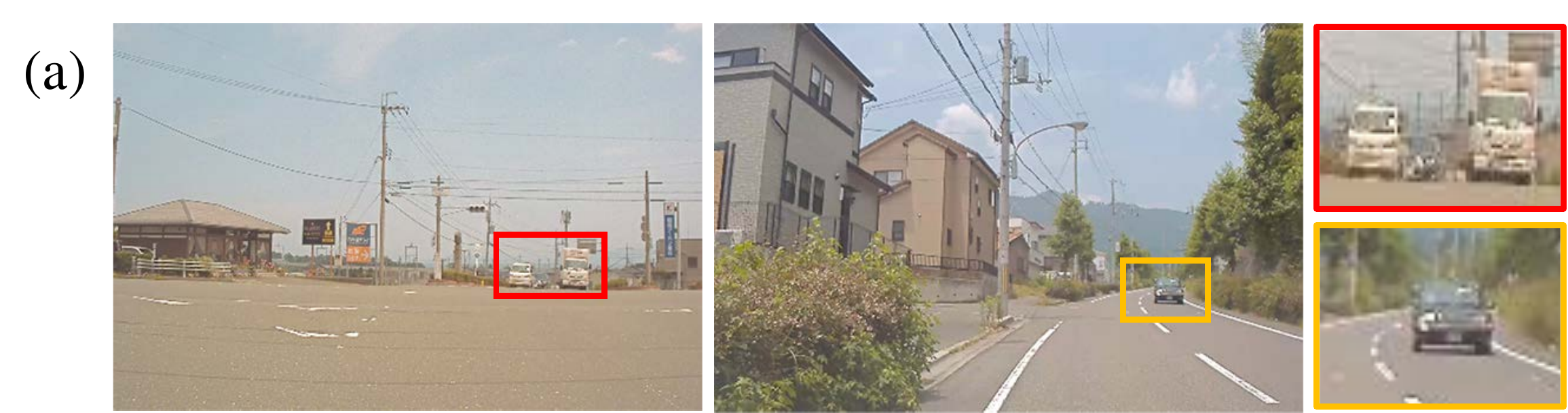}\\ \vspace{0.5pt}
\includegraphics[width=1\linewidth]{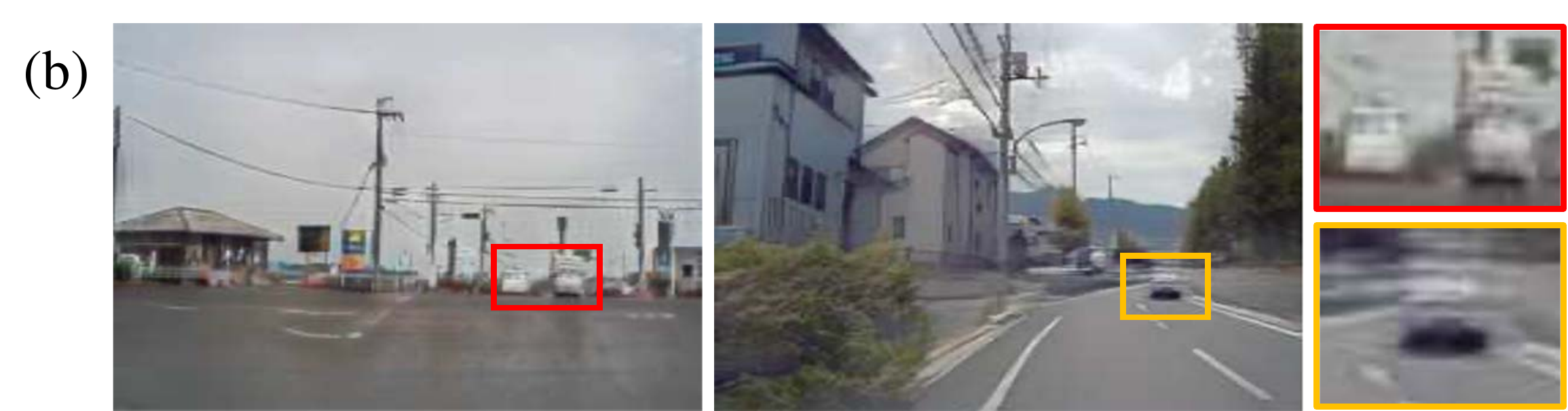}\\  \vspace{0.5pt}
\includegraphics[width=1\linewidth]{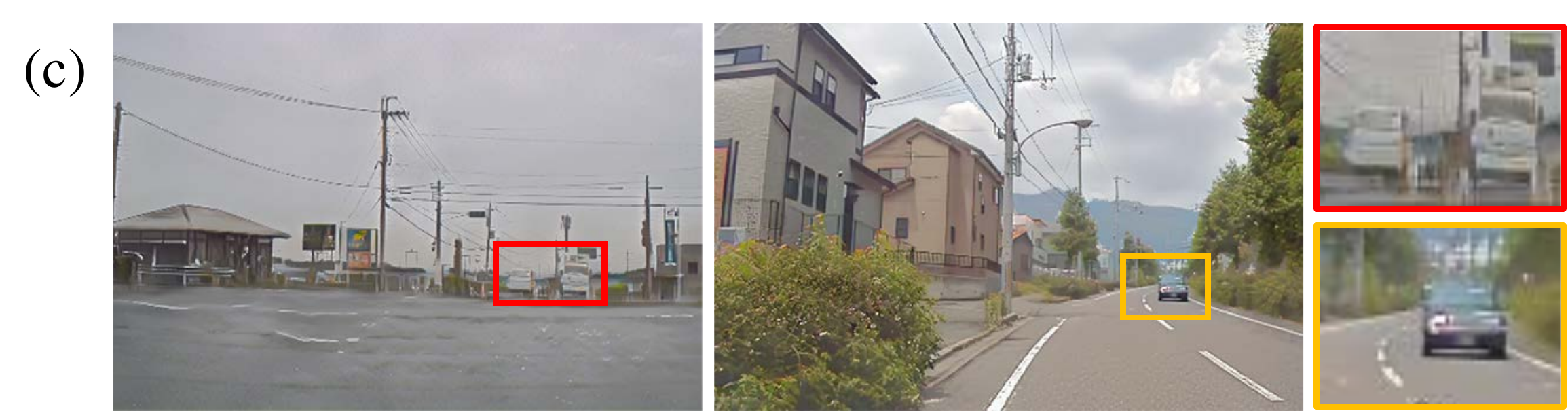}\\ \vspace{-4pt}
\caption{\textbf{Visual comparison of DUNIT~\cite{bhattacharjee2020dunit}.}
(a) Input, (b) DUNIT, (c) Ours. We show the results for sunny$\rightarrow$rainy in the first column and sunny$\rightarrow$cloudy in the second column. Note that the results are taken from DUNIT paper.}
\label{fig:f5} \vspace{-5pt}
\end{figure}

\begin{figure}[t]
\includegraphics[width=0.247\linewidth]{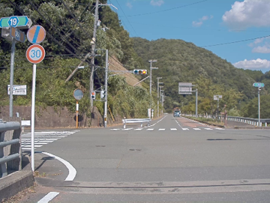}\hfill
\includegraphics[width=0.247\linewidth]{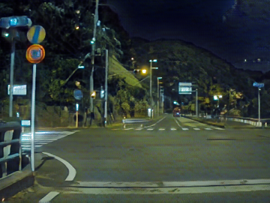}\hfill
\includegraphics[width=0.247\linewidth]{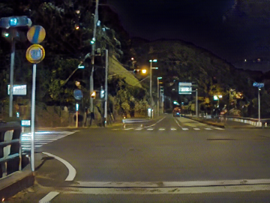}\hfill
\includegraphics[width=0.247\linewidth]{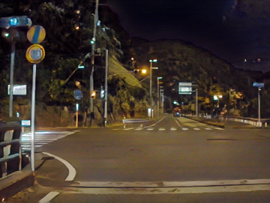}\hfill \\
\includegraphics[width=0.247\linewidth]{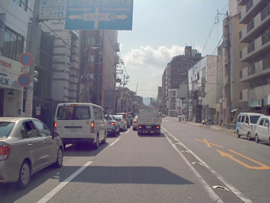}\hfill
\includegraphics[width=0.247\linewidth]{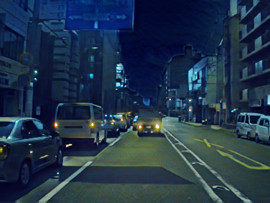}\hfill
\includegraphics[width=0.247\linewidth]{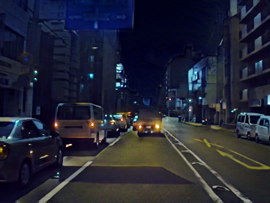}\hfill
\includegraphics[width=0.247\linewidth]{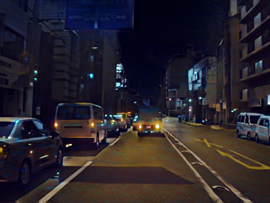}\hfill \\
\vspace{-18pt}
\caption{\textbf{Results of multimodal image translation.}
We use randomly sampled style values to generate (left) sunny image $\rightarrow$ (right) night images.}
\vspace{-5pt}
\label{fig:multi}
\end{figure}

\vspace{-5pt}
\paragraph{Qualitative evaluation.}
Fig.~\ref{fig:f4} shows a qualitative comparison of the state-of-the-art methods.
We observe that the multi-modal I2I methods MUNIT~\cite{huang2018multimodal} and DRIT~\cite{lee2018diverse} fails to capture instance details and boundaries well.
As these methods do not have any access to semantic information, they tend to translate instances to the other semantic styles (e.g., translating buildings into the sky).
Our method produces the most visually appealing images with more vivid details.
Thanks to the proposed class-aware memory network, it shows high capacity to better understand the semantic instances and employ the appropriate local style representation for object classes.
We compare the translated results with instance-level I2I method DUNIT~\cite{bhattacharjee2020dunit} in Fig.~\ref{fig:f5}.
Our result yields sharper and distinctive instances and more realistic images.
Lastly, we visualize the multimodal translated results in Fig.~\ref{fig:multi}.
We use the stored key $\mathbf{k}$ in the memory and randomly sampled values $(\mathbf{v}^x, \mathbf{v}^y)$.
It can be observed that the degree of color (\eg road, sky) changes across these images.

\begin{table*}[]
\centering
\resizebox{\textwidth}{!}{
\begin{tabular}{l|cc|cc|cc|cc|cc|cc|cc}
\hline
& \multicolumn{2}{c|}{CycleGAN~\cite{zhu2017unpaired}} & \multicolumn{2}{c|}{UNIT~\cite{liu2017unsupervised}} & \multicolumn{2}{c|}{MUNIT~\cite{huang2018multimodal}} & \multicolumn{2}{c|}{DRIT~\cite{lee2018diverse}} & \multicolumn{2}{c|}{INIT~\cite{shen2019towards}} & \multicolumn{2}{c|}{DUNIT~\cite{bhattacharjee2020dunit}} & \multicolumn{2}{c}{Ours} \\ \cline{2-15}
& CIS & IS & CIS & IS & CIS & IS & CIS & IS & CIS & IS & CIS & IS & CIS & IS \\ \hline\hline
sunny$\rightarrow$night	&	0.014	&	1.026	&	0.082	&	1.030	&	1.159	&	1.278	&	1.058	&	1.224	&	1.060	&	1.118	&	1.166	&	1.259	&	1.176	&	1.271	\\
night$\rightarrow$sunny	&	0.012	&	1.023	&	0.027	&	1.024	&	1.036	&	1.051	&	1.024	&	1.099	&	1.045	&	1.080	&	1.083	&	1.108	&	1.115	&	1.130	\\\hline
sunny$\rightarrow$rainy	&	0.011	&	1.073	&	0.097	&	1.075	&	1.012	&	1.146	&	1.007	&	1.207	&	1.036	&	1.152	&	1.029	&	1.225	&	1.092	&	1.213	\\
sunny$\rightarrow$cloudy	&	0.014	&	1.097	&	0.081	&	1.134	&	1.008	&	1.095	&	1.025	&	1.104	&	1.040	&	1.142	&	1.033	&	1.149	&	1.052	&	1.218	\\
cloudy$\rightarrow$sunny	&	0.090	&	1.033	&	0.219	&	1.046	&	1.026	&	1.321	&	1.046	&	1.249	&	1.016	&	1.460	&	1.077	&	1.472	&	1.136	&	1.489	\\\hline\hline
Average	&	0.025	&	1.057	&	0.087	&	1.055	&	1.032	&	1.166	&	1.031	&	1.164	&	1.043	&	1.179	&	1.079	&	1.223	&	\textbf{1.112}	&	\textbf{1.254}	\\\hline
\end{tabular}}
\vspace{-5pt}
\caption{\textbf{Quantitative evaluation on INIT dataset~\cite{shen2019towards}.}
We perform bidirectional translation for each domain pair.
We measure CIS and IS (higher is better).
Our results attain the best results.}
\vspace{-13pt}
\label{tab1}
\end{table*}
\vspace{-3pt}

\begin{figure}[]
	\centering
    \vspace{-5pt}
	\renewcommand{\thesubfigure}{}
	\subfigure[(a) Image quality]
	{\includegraphics[height=2.55cm]{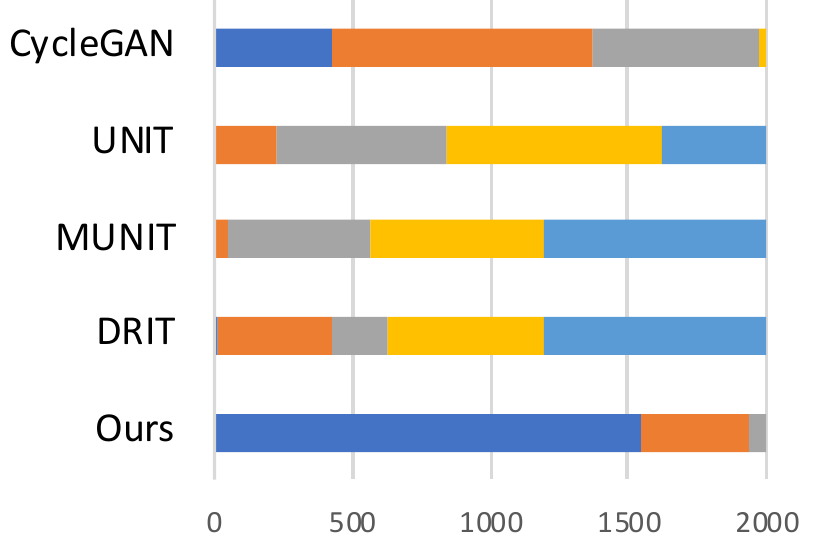}}\hfill
	\subfigure[(b) Style diversity]
	{\includegraphics[height=2.55cm]{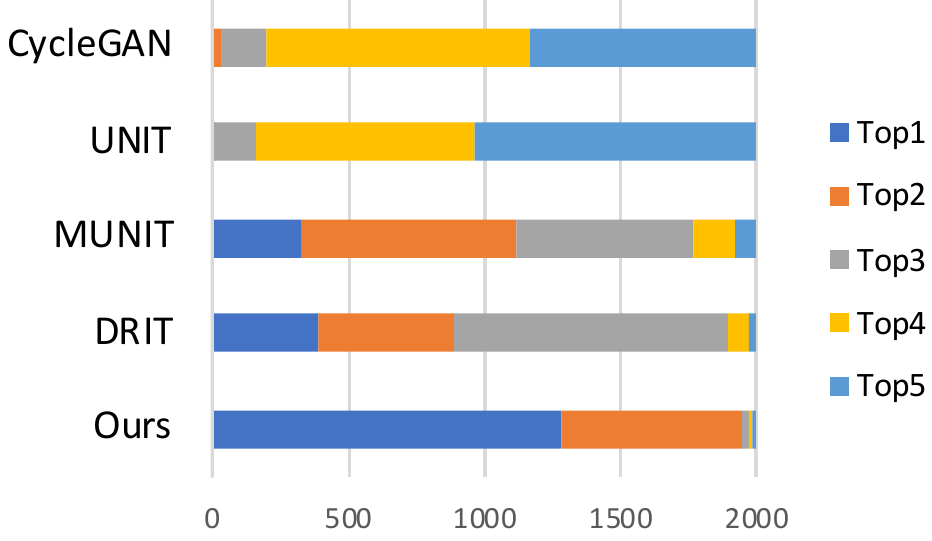}}\hfill \\ \vspace{-10pt}
	\caption{\textbf{User study results.} Our method is most preferred for image quality and style diversity both.}
	\vspace{-4pt}
	\label{fig:f11}
\end{figure}

\vspace{-12pt}
\paragraph{User study.}
We conducted a user study to compare subjective quality of the translated results. 
For each translation, we randomly select 10 images from INIT validation to set up a total of 80 images for comparison.
From 25 participants, we asked to rank all the methods in terms of the image quality and style diversity of the translated image, and we received a total of 2,000 votes.
Fig.~\ref{fig:f11} shows the results, and our method ranks first in 77.9$\%$ for the image quality and 64.5$\%$ for the style diversity.

\vspace{-12pt}
\paragraph{Quantitative evaluation.}
Table ~\ref{tab1} shows the IS~\cite{salimans2016improved} and  CIS~\cite{huang2018multimodal}, and Table~\ref{tab2} shows average LPIPS metric~\cite{zhang2018unreasonable}.
The IS measures the diversity of output images based on the Inception V3 model~\cite{szegedy2016rethinking}.
The CIS quantifies the quality and diversity of output conditioned on a single image.
Additionally, the LPIPS metric~\cite{zhang2018unreasonable} measures the translation diversity by calculating the similarity between two different deep features from the pre-trained AlexNet~\cite{krizhevsky2017imagenet}.
The results indicate significant performance gains with our method in all metrics.
It further highlights the contribution of class-aware memory network to the improved performance.

\begin{table}[]
\centering
\resizebox{1\linewidth}{!}{
\begin{tabular}{l|c|c|c|c}
\hline
\multirow{2}{*}{Method}	&	sunny	&	sunny	&	sunny	& \multirow{2}{*}{Average}		\\ 
	&	$\rightarrow$night	&	$\rightarrow$rainy	&	$\rightarrow$cloudy	&
\\\hline\hline
CycleGAN~\cite{zhu2017unpaired}	&	0.016	&	0.008	&	0.011	&	0.012	\\
UNIT~\cite{liu2017unsupervised}	&	0.067	&	0.062	&	0.068	&	0.066	\\
MUNIT~\cite{huang2018multimodal}	&	0.292	&	0.239	&	0.211	&	0.247	\\
DRIT~\cite{lee2018diverse}	&	0.231	&	0.173	&	0.166	&	0.190	\\
INIT~\cite{shen2019towards}	&	0.330	&	0.267	&	0.224	&	0.274	\\
DUNIT~\cite{bhattacharjee2020dunit}	&	0.338	&	0.298	&	0.225	&	0.287	\\
Ours	&	\textbf{0.346}	&	\textbf{0.316}	&	0\textbf{.251}	&	\textbf{0.304}	\\\hline
Real images	&	0.573	&	0.489	&	0.465	&	0.509	\\\hline
\end{tabular}}
\vspace{-5pt}
\caption{\textbf{Quantitative evaluation with average LPIPS metric.}
The LPIPS metric calculates the diversity scores.}
\label{tab2}
\vspace{-2pt}
\end{table}

\begin{table}[]
\centering
\resizebox{0.95\linewidth}{!}{
\begin{tabular}{l|c|c|c|c|c}
\hline
\multirow{2}{*}{Method}&\multicolumn{5}{c}{KITTI $\rightarrow$ Cityscapes}\\
\cline{2-6}
	&	Pers.	&	Car	&	Truc.	&	Bic.	&	\textbf{mAP}	\\ \hline\hline
DT~\cite{inoue2018cross}	&	28.5	&	40.7	&	25.9	&	29.7	&	31.2	\\
DAF~\cite{chen2018domain}	&	39.2	&	40.2	&	25.7	&	48.9	&	38.5	\\
DARL~\cite{kim2019diversify}	&	46.4	&	58.7	&	27.0	&	49.1	&	45.3	\\
DAOD~\cite{rodriguez2019domain}	&	47.3	&	59.1	&	28.3	&	49.6	&	46.1	\\
DUNIT w/o IC~\cite{bhattacharjee2020dunit} &	56.2	&	59.5	&	24.9	&	48.2	&	47.2	\\
DUNIT w/ IC~\cite{bhattacharjee2020dunit} &	\textbf{60.7}	&	65.1	&	32.7	&	57.7	&	54.1	\\
Ours&	58.3	&	\textbf{68.2}	&	\textbf{33.4}	& \textbf{58.4}	&	\textbf{54.6}\\ \hline
\end{tabular}}
\vspace{-5pt}
\caption{\textbf{Quantitative results for domain adaptive detection.}
We report per-class AP for KITTI$\rightarrow$Cityscapes case.}
\vspace{-10pt}
\label{tab3}
\end{table}

\begin{figure*}[] 
	\centering
	\renewcommand{\thesubfigure}{}
    {\includegraphics[width=0.198\linewidth]{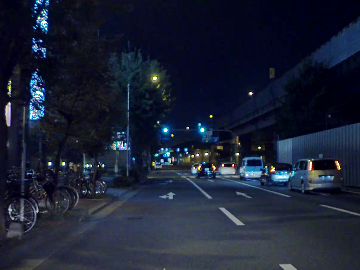}}\hfill
	{\includegraphics[width=0.198\linewidth]{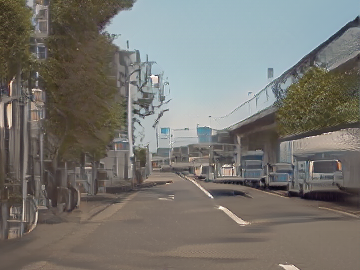}}\hfill
	{\includegraphics[width=0.198\linewidth]{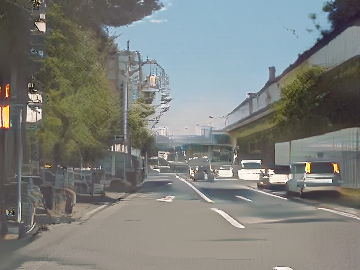}}\hfill
	{\includegraphics[width=0.198\linewidth]{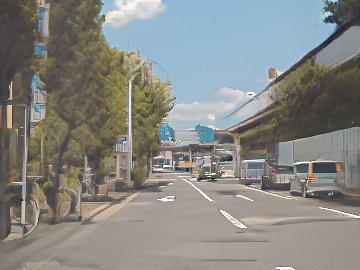}}\hfill
	{\includegraphics[width=0.198\linewidth]{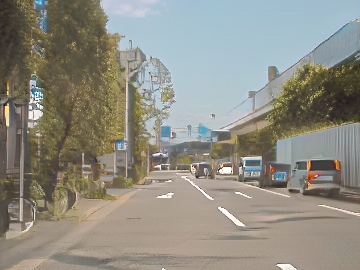}}\hfill \\ \vspace{-5pt}
    \subfigure[(a) Input]
	{\includegraphics[width=0.198\linewidth]{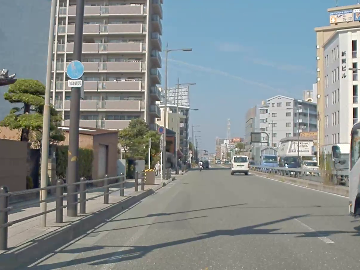}}\hfill
	\subfigure[(b) w/ sm+tl]
	{\includegraphics[width=0.198\linewidth]{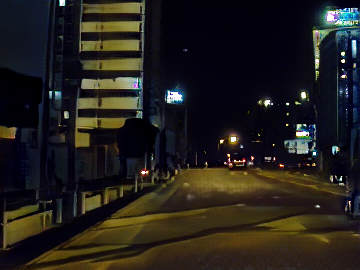}}\hfill
    \subfigure[(c) w/ sm+cl]
	{\includegraphics[width=0.198\linewidth]{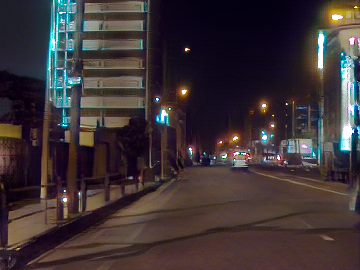}}\hfill
	\subfigure[(d) w/ cm+tl]
	{\includegraphics[width=0.198\linewidth]{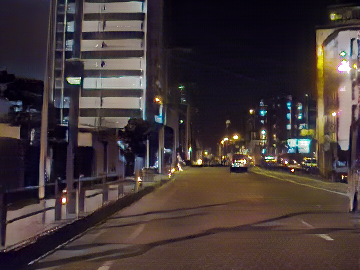}}\hfill
	\subfigure[(e) w/ cm+cl]
	{\includegraphics[width=0.198\linewidth]{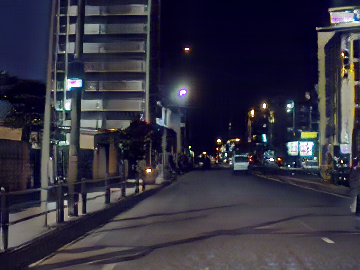}}\hfill \\ \vspace{-10pt}
	\caption{\textbf{Qualitative evaluation for ablation study.}
    Our full configuration with class-aware memory and contrastive loss produces a realistic and well-preserved image.}\vspace{-12pt}
	\label{fig:f9}
\end{figure*}

\begin{figure}[]
\vspace{-4pt}
	\centering
	\subfigure[w/ cm+tl]
	{\includegraphics[width=0.46\linewidth]{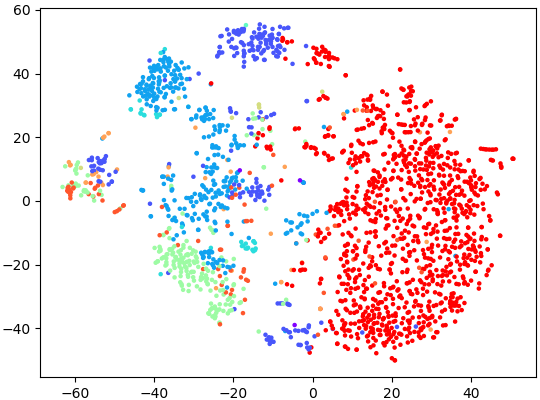}}\hfill
	\subfigure[w/ cm+cl]
	{\includegraphics[width=0.46\linewidth]{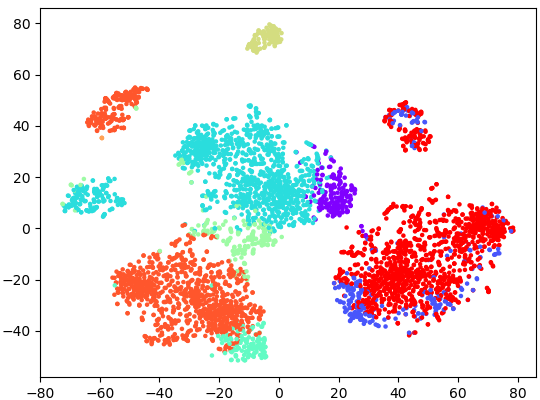}}\hfill \\
    \vspace{-10pt}
	\caption{\textbf{t-SNE visualization for the content features.} The same colored points indicate the content features addressed to the same memory item.}
	\label{fig:f7}\vspace{-10pt}
\end{figure}

\vspace{-12pt}
\paragraph{Domain adaptation for object detection.}
We test our method for the domain adaptive object detection.
Using the Faster-RCNN~\cite{ren2016faster} trained on images in the source domain, we evaluate the detection performance of the translated images from source to target domain.
Following DUNIT~\cite{bhattacharjee2020dunit}, we conduct experiments on the KITTI object detection benchmark~\cite{geiger2012we} as the source domain and Cityscapes dataset~\cite{cordts2016cityscapes} as the target domain.
We compare the performance to state-of-the-art domain adaptation methods~\cite{inoue2018cross,chen2018domain,kim2019diversify,rodriguez2019domain} and DUNIT~\cite{bhattacharjee2020dunit} with instance consistency loss (w/ IC) and without (w/o IC).
Note that the instance consistency loss enforces the consistency constraints between results detected from original and translated image.
We report the mean average precision (mAP) for the detected objects in Tab.~\ref{tab3}.
Our method performed well in the domain adaptive object detection tasks without explicitly using the object detection network.
Unlike DUNIT~\cite{bhattacharjee2020dunit} that improves performance by applying direct constraints on detected results, 
our method can recognize the semantic information contained in images thanks to our highly discriminative class-aware memory network.
Consequently, it allows the image that are translated into an appropriate object style while preserving its inherent semantic information.
Furthermore, it demonstrates that our method can realize more complex domain adaptation tasks.

\vspace{-3pt}
\subsection{Ablation study}
\vspace{-3pt}
We examine the impact of i) single memory (\emph{sm}) without considering object class vs. class-aware memory (\emph{cm}) and ii) feature triplet loss (\emph{tl}) vs. feature contrastive loss (\emph{cl}).
We conduct the ablation studies on sunny $\leftrightarrow$ night case in INIT~\cite{shen2019towards}, which is apt to show the effectiveness of individual components.
We compare the results of  4 cases; (a) w/ single memory + triplet loss, (b) w/ single memory + contrastive loss, (c) w/ class-aware memory + triplet loss, and (d) w/ class-aware memory + contrastive loss.
The qualitative and quantitative results are shown in Fig.~\ref{fig:f9} and Table~\ref{tab4}.

\vspace{-12pt}
\paragraph{Effectiveness of class-aware memory.}
The results using the single memory (in Figure~\ref{fig:f9} (b), (c)) cannot preserve the instance boundaries well, and even small instances disappear into the background.
On the other hand, the results using the class-aware memory (in Figure~\ref{fig:f9} (d), (e)) show clear and well-preserved instance structures.
The quantitative results from Table~\ref{tab4} also indicate that the translated images using the class-aware memory are more realistic.

\paragraph{Effectiveness of feature contrastive loss.}
We observe that the results using the feature contrastive loss (in Figure~\ref{fig:f9} (c), (e)) are more vivid and represent a style that is appropriate for each instance compared to the results using the feature triplet loss (in  Figure~\ref{fig:f9} (b), (d)).
To investigate its effect, we visualize the distribution of the content features, which are learned with the triplet loss in Fig.~\ref{fig:f7} (a) and with the contrastive loss in Fig.~\ref{fig:f7} (b).
Specifically, we project the embedded content features from the test images into 2-dimensional space using t-SNE~\cite{maaten2008visualizing}.
The color indicates the memory items, which means that the points with the same color are mapped to the same item.
The contrastive loss is more effective in separating and clustering the feature semantically.
Therefore, it enhances the diversity and discriminative power of our memory items.

\begin{table}[]
\centering
\resizebox{0.88\linewidth}{!}{
\begin{tabular}{l|ccc|cc}
\hline
\multirow{2}{*}{Method}	&	\multicolumn{3}{c|}{sunny$\rightarrow$night}	&\multicolumn{2}{c}{night$\rightarrow$sunny}\\ \cline{2-6}
& LPIPS & CIS & IS & CIS & IS \\ \hline\hline
w/ sm+tl	& 0.287 & 1.061 & 1.189 & 1.037 & 1.080 \\
w/ sm+cl	& 0.310 & 1.094 & 1.206 & 1.062 & 1.107 \\
w/ cm+tl	& 0.328 & 1.156 & 1.253 & 1.101 & 1.103 \\
w/ cm+cl  & \textbf{0.346} & \textbf{1.176} & \textbf{1.271} &\textbf{1.115} & \textbf{1.130} \\
\hline
\end{tabular}}\vspace{-5pt}
\caption{\textbf{Ablation study on memory types and memory losses.}
Our full configuration shows the best performance.
}\vspace{-8pt}
\label{tab4}
\end{table}

\vspace{-4pt}
\section{Conclusion}
\vspace{-3pt}
We present an instance-level unsupervised image-to-image translation framework with a class-aware memory network.
It consists of a set of key-values that store shared content and domain-specific style representations, used to explicitly reason style representations.
To this end, we introduce feature contrastive loss to increase the diversity and discriminative power of our memory items.
This allows obtaining object-preserved and high-quality translated outputs without the additional use of extra object detection modules.
Extensive experiments show that our method achieves state-of-the-art performance.

{\small
\bibliographystyle{ieee_fullname}
\bibliography{egbib.bbl}
}

\end{document}